\documentclass[10pt,twocolumn,letterpaper]{article}

\usepackage[pagenumbers]{cvpr} 

\usepackage[dvipsnames]{xcolor}

\definecolor{cvprblue}{rgb}{0.21,0.49,0.74}
\usepackage[pagebackref,breaklinks,colorlinks,citecolor=cvprblue]{hyperref}
\usepackage{algorithm}
\usepackage{algorithmic}
\usepackage{multirow}
\usepackage{colortbl}
\usepackage{mathrsfs}

\def\paperID{***} 
\def\confName{3DV\xspace}
\def\confYear{2025\xspace}

\title{G-NeLF: Memory- and Data-Efficient Hybrid Neural Light Field for Novel View Synthesis}

\author{Lutao Jiang$^{1}$ \quad Lin Wang$^{1}$$^{,2}$ \footnotemark$^{*}$\\
$^{1}$ VLIS LAB, AI Thrust, HKUST(GZ) \quad $^{2}$Dept. of CSE, HKUST
\\
{\tt\small ljiang553@connect.hkust-gz.edu.cn, linwang@ust.hk} \\
\small{Project Page: \url{https://vlislab22.github.io/G-NeLF/}}
}
\begin{document}

\twocolumn[{
\renewcommand\twocolumn[1][]{#1}%
\maketitle
\begin{center}
\centering
\vspace{-20pt}
  \includegraphics[width=\textwidth]{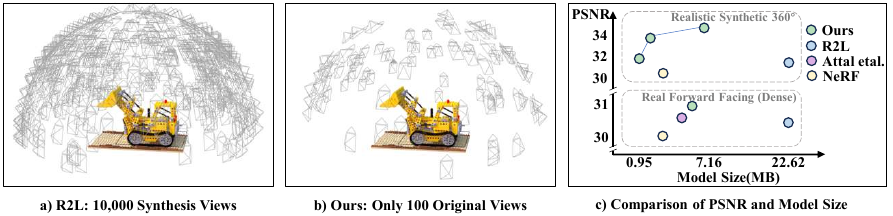}
\captionsetup{font=small}
\end{center}
\vspace{-15pt}
\captionof{figure}{\textbf{A comparison between R2L~\cite{wang2022r2l} and our G-NeLF}. a) R2L needs to use a pre-trained NeRF model to synthesize 10,000 training images to train the NeLF model. b) Our G-NeLF only needs the original amount of images the same as NeRF's dataset. c) The comparison of rendering quality and model size among ours, R2L, and NeLF proposed by Attal \etal \cite{attal2022learning}. The upper half is the Realistic Synthetic 360° dataset. The lower half is scenes containing relatively dense views in the Real Forward-Facing Dataset.}
\vspace{5pt}
\label{fig:1}
}]
\maketitle

\renewcommand{\thefootnote}{\fnsymbol{footnote}}
\footnotetext[1]{Corresponding author.}
\renewcommand{\thefootnote}{\arabic{footnote}}

\begin{abstract}
\vspace{-10pt}
Following the burgeoning interest in implicit neural representation, Neural Light Field (NeLF) has been introduced to predict the color of a ray directly.
Unlike Neural Radiance Field (NeRF), NeLF does not create a point-wise representation by predicting color and volume density for each point in space.
However, the current NeLF methods face a challenge as they need to train a NeRF model first and then synthesize over 10\textit{K} views to train NeLF for improved performance.
Additionally, the rendering quality of NeLF methods is lower compared to NeRF methods.
In this paper, we propose \textbf{G-NeLF}, a versatile grid-based NeLF approach that utilizes spatial-aware features to unleash the potential of the neural network's \textbf{inference capability}, and consequently overcome the difficulties of NeLF training.
Specifically, we employ a spatial-aware feature sequence derived from a meticulously crafted grid as the ray's representation.
Drawing from our empirical studies on the adaptability of multi-resolution hash tables, we introduce a novel grid-based ray representation for NeLF that can represent the entire space with a very limited number of parameters. 
To better utilize the sequence feature, we design a lightweight ray color decoder that simulates the ray propagation process, enabling a more efficient inference of the ray's color.
G-NeLF can be trained \textbf{without necessitating significant storage overhead} and with \textbf{the model size of only 0.95 MB} to surpass previous state-of-the-art NeLF.
Moreover, compared with grid-based NeRF methods, \eg, Instant-NGP, we only utilize one-tenth of its parameters to achieve higher performance.
Our code will be released upon acceptance.
\end{abstract}

\vspace{-15pt}
\section{Introduction}

Driven by inspiration from Implicit Neural Representation (INR) \cite{park2019deepsdf, mescheder2019occupancy}, the task of novel view synthesis -- representing a 3D scene from 2D views -- has recently witnessed a significant breakthrough with the introduction of Neural Radiance Fields (NeRF)~\cite{mildenhall2020nerf, wang2021nerf, zhang2020nerf++, martin2021nerf, barron2021mip, chen2022tensorf, muller2022instant, garbin2021fastnerf, wang20224k, yang2022ps, radl2023analyzing, li2023compressing, chen2023mobilenerf, gao2023strivec, hu2023tri, zheng2023multi} and Neural Light Field (NeLF)~\cite{sitzmann2021light, suhail2022light, attal2022learning, wang2022r2l, cao2023real, yu2023dylin, gupta2024lightspeed}.
\textit{A key distinction between NeRF and NeLF is the necessity for the physical modeling of each point.}
NeRF requires querying the color and volume density of each point in the space from a neural network and then utilizing the classical volume rendering~\cite{kajiya1984ray} to render an image.
In contrast, the NeLF network predicts the ray's color directly based on the ray coordinate, \eg, the coordinates of two points can encode a ray's coordinate.
However, since there are no constraints based on physical world modeling, it is difficult for the neural network to learn the color of each ray.

A seminal work, R2L~\cite{wang2022r2l}, proposes a deep MLP network to represent the light field faithfully by leveraging the network's greater \textit{memory capability} and expressivity. Unfortunately, this causes difficulties in training such a cumbersome network that only uses 2D images with 100 views (commonly used for training NeRF). For this reason, it synthesizes 10K images with various viewpoints using a pre-trained NeRF.
The following NeLF methods~\cite{cao2023real, gupta2024lightspeed} also adhere to this design.
However, such an image synthesis operation takes a long time and consumes storage for the record rays' coordinates and color. This renders challenges for applications that require images with higher resolutions.
Consequently, \textit{it remains challenges of how to reduce computation cost and alleviate the training difficulty of NeLF under the condition of data shortage}.

In this paper, we present \textbf{G-NeLF}, a lightweight NeLF approach hybridized with spatial-aware grids, which overcomes the training difficulties of R2L with only 100 views or fewer for training, as demonstrated in the comparison in Fig.\ref{fig:1} (a) and (b).
Also, G-NeLF can be trained without requiring significant storage overhead with the model size of only \textbf{0.95 MB}, see Fig.~\ref{fig:1}(c). \textit{Our key idea} is to utilize spatial-aware feature sequence as the ray representation method to unleash the potential of the recurrent network's \textit{inference capability} by leveraging the \textit{temporal order information} in ray propagation.

To obtain the feature sequence, we design a 3D spatial-aware feature grid in the space.
Instant-NGP \cite{muller2022instant} introduces a multi-resolution hash table designed to store features of 3D grids.
While the hash table reduces the parameter count compared with the vanilla 3D grid, its quantity remains relatively large.
Our analysis and empirical evidence (See Tab.~\ref{tab: Hash Grid} and Fig.~\ref{fig:Hash Grid}) reveal that in Instant-NGP's hash table, low-resolution features primarily encode shape information, while high-resolution features predominantly capture details related to surface appearance and color.
We analyze it in Sec.~\ref{sec: Grid-based Ray Representation}.
Though we don't model the point-wise volume density representation, we can still assimilate this property into our architecture by the integration of multiple hash-based multi-resolution grids, benefiting from the \textit{spatial awareness of our ray representation}. 
Based on this, we present a 3D grid decomposition representation for NeLF, which only comprises a compact set of 474,432 parameters.
Given that ray propagation inherently possesses strong temporal order information, efficiently capitalizing on this aspect is crucial to achieving our objectives.
Therefore, we design an LSTM-based~\cite{hochreiter1997long} ray color decoder (Sec.~\ref{sec: Ray Color Decoder}), a structure that experiments suggested is optimal, to predict the ray's color throughout the sequence.

Our contribution can be summarized as follows:
\begin{itemize}
    \item We present G-NeLF, an \textit{implicit} approach for the novel view synthesis task, \ie, not depend on intricate point-wise volume-based physical modeling techniques commonly used in NeRFs, and \textit{not depend on other auxiliary data~\cite{wang2022r2l} for training}.

    \item We propose a novel grid-based ray representation method, which can unleash the potential of \textit{inference ability} of the neural networks. Also, we design a lightweight ray color decoder, which simulates the temporal nature of ray propagation and is extremely \textit{flexible} to adjust the model size.

    \item Drawing on the analysis of the integration between multi-resolution and hash mapping, we propose a novel 3D grid representation for NeLF characterized by an \textit{exceptionally small number of parameters}.

    \item We conduct extensive experiments to prove the superiority of our G-NeLF. As depicted in Fig.\ref{fig:1} (c), compared with SoTA NeLF methods, \eg, R2L~\cite{wang2022r2l}, we surpass it by \textbf{0.04dB} only with \textbf{100} views and \textbf{0.95} MB model size, and surpass it by \textbf{2.48dB} with the increase of model size; Compared with SoTA lightweight NeRF compression method, VQRF~\cite{li2023compressing}, we use same model size, \textbf{1.4MB}, to surpass it by \textbf{0.79dB};
    Compared with NeRF methods~\cite{barron2021mip, chen2022tensorf, muller2022instant}, we achieve comparable results.
\end{itemize}

\vspace{-5pt}
\section{Related Works}
\vspace{-2pt}

\begin{figure*}[t!]
    \centering
    \includegraphics[width=1\linewidth]{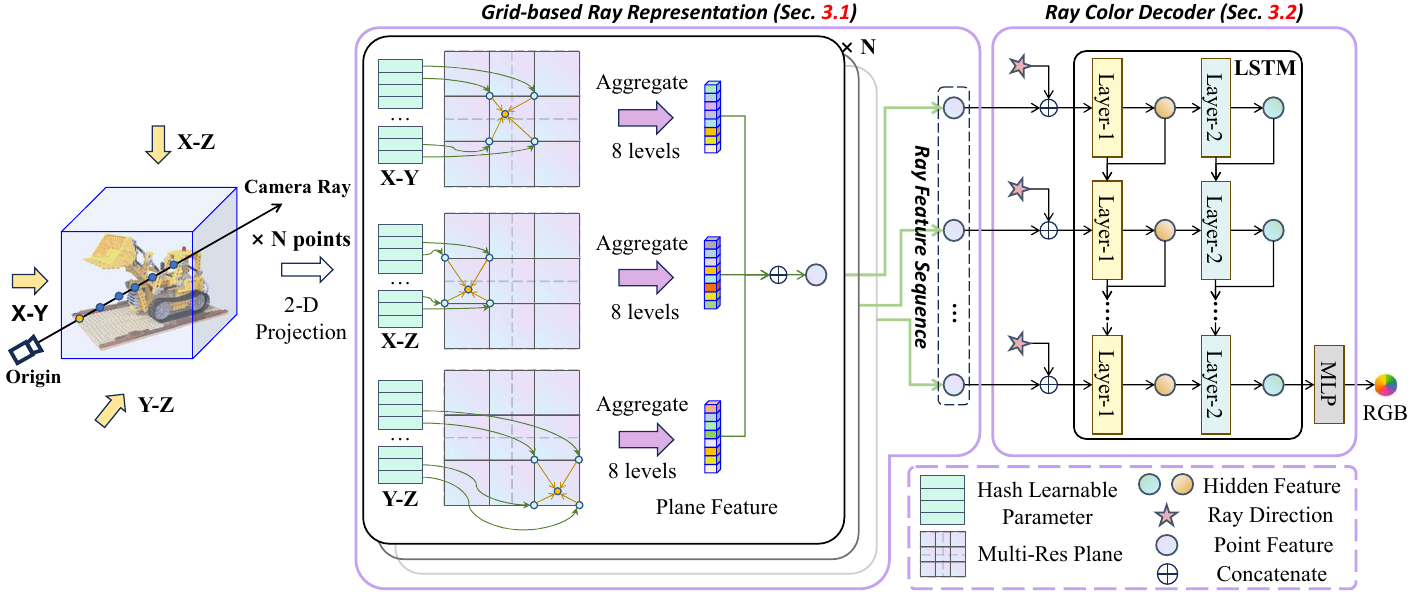}
    \vspace{-20pt}
    \caption{\textbf{An overview of our G-NeLF framework}. For one ray emitted from a camera, we first orderly sample a few points on it and use our designed hash multi-resolution triplane to obtain their feature. Then these features will compose the ray's feature sequence to represent this ray. Finally, we design a ray color decoder to transform the ray feature sequence into RGB color.}
    \vspace{-13pt}
    \label{fig:Framework}
\end{figure*}

\noindent\textbf{Neural Light Field.}
The concept of the light field was initially introduced by \cite{Levoy_Hanrahan_1996, Gortler_Grzeszczuk_Szeliski_Cohen_1996}.
Both studies employ a 4-D coordinate system ($(u,v)$, $(s, t)$) to represent a ray within space.
By pinpointing the positions of $(u,v)$ and $(s, t)$ on two distinct planes, the ray is consequently determined.
The recent advancements in NeRF \cite{mildenhall2020nerf} have catalyzed interest in Implicit Neural Representation (INR). 
Consequently, an increasing number of researchers are recognizing the potential of INR in the domain of the light field.
Light Field Networks (LFNs) \cite{sitzmann2021light} pioneers the integration of INR with the light field. 
They utilize meta-learning to get the prior of multi-view consistent light field and use one single view image to reconstruct the other views.
But compared with NeRFs, it remains a relatively low quality.
Suhail \etal \cite{suhail2022light} utilize transformer \cite{vaswani2017attention} structure to factor in epipolar lines from various views for novel views. 
But its inference speed and training cost is unacceptable.
Attal \etal \cite{attal2022learning} propose to map 4D ray to an embedding space.
They use some voxels to represent the local light field while simultaneously learning the opacity of the voxels.
R2L \cite{wang2022r2l} introduces the innovative method of using $K$ points, as opposed to the conventional two points, to represent a ray.
To mitigate the data constraints inherent in NeLF training, they synthesize a lot of pseudo data from the pre-trained NeRF model.
DyLiN \cite{yu2023dylin} first makes the light field dynamic depending on a deformation network.
MobileR2L \cite{cao2023real} incorporates a $1\times1$ convolution layer as a substitute for the MLP used in R2L, and integrates multiple lightweight super-resolution modules to further reduce the computation consumption, which makes it run on iPhone.
LightSpeed~\cite{gupta2024lightspeed} adopts the K-Planes~\cite{fridovich2023k} approach, employing six feature planes to encode the $4$-D two-plane coordinates of rays, while simultaneously integrating a super-resolution module~\cite{cao2023real} to increase the resolution.
However, compared with R2L and MobileR2L, LightSpeed imposes an increasing demand for the training dataset of 30$K$ pseudo images.

Differently, we harness spatial information to unleash the potential of the inference ability of neural networks. This solves the problems of previous works, including training data requirements and the complexity of the whole training pipeline.
Moreover, we only use the original training data with a 0.95 MB model size to surpass R2L.

\noindent\textbf{Grid Design in NeRF.}
NeRF~\cite{mildenhall2020nerf} is proposed for the novel view synthesis task and 3D reconstruction task, and it has become the dominant paradigm.
However, the low training and inference speed present challenges for real-world applications.
Recently, a lot of methods are proposed to solve these problems mainly from point sampling aspect~\cite{liu2020neural, lindell2021autoint, jiang2023sdf} and data structure aspect including OcTree~\cite{yu2021plenoctrees, wang2022fourier, bai2023dynamic}, mesh~\cite{chen2023mobilenerf,tang2023delicate} and grid representation~\cite{garbin2021fastnerf, muller2022instant, wang2022neus2, chen2022tensorf, tang2022compressible, gao2023iCCV, fridovich2023k, cao2023hexplane, takikawa2022variable, li2023compressing}.
Instant-NGP~\cite{muller2022instant} represents the most advanced fusion of NeRF and explicit grid representation, which maps the multi-resolution grid to a table using hash function.
Although the hash table has reduced the number of parameters than the naive 3D grid, the amount is still too large. 
EG3D~\cite{chan2022efficient} introduces the tri-plane to represent a NeRF Grid.
TensoRF~\cite{chen2022tensorf} proposes a 3D decomposition method that uses three planes and three axes to represent a 3D grid.
However, due to its expansive feature dimension width, the amount of parameters is still too large.
VQRF~\cite{li2023compressing} uses vector quantization and weight quantization techniques to compress the Grid to 1MB.
K-Planes~\cite{fridovich2023k} and HexPlane~\cite{cao2023hexplane} extend the grid design to dynamic scenes by composing several 2D planes of X, Y, Z, and T axes.

Different from these methods, we delve deeper into the role of the hash mapping function during the training phase and demonstrate its adaptability and flexibility.
Building on this foundation, we formulate a highly efficient grid representation for NeLF, requiring only about 0.47 million parameters to surpass the previous methods, \eg, R2L.
While NeRF methods typically demand more sampling points to ensure precise volume density modeling and spatial stability, our G-NeLF simplifies the process. 
Our G-NeLF only needs to sample fewer points' features to represent one ray's feature, avoiding more complex training requirements.

\vspace{-5pt}
\section{Method}
\vspace{-2pt}

\noindent \textbf{Overview.}
The goal of G-NeLF is to synthesize arbitrary views of the scene.
The design is guided by three key ideas.
\textbf{1)} We employ a feature sequence as a substitute for the previous coordinate-based ray representation.
\textbf{2)} Based on our analysis, we design a hash-based multi-resolution tri-plane.
\textbf{3)} We rely on the strong temporal information embedded within the ray's feature sequence to infer its color.

Our workflow, as depicted in Fig.~\ref{fig:Framework}, encompasses the following steps.
We emit rays based on the camera model to form an image, with each ray corresponding to a pixel.
Initially, the ray crosses a bounding box in space.
Next, we sequentially sample some points on the ray and project them onto three orthogonal feature planes.
Utilizing the feature grid we designed, we obtain the feature vector for each point and systematically assemble these vectors to form the ray's feature sequence (Sec.~\ref{sec: Grid-based Ray Representation}).
To decode the ray's color, we design a lightweight ray color decoder based on an LSTM network (Sec.~\ref{sec: Ray Color Decoder}).
This decoder receives the ray's feature sequence along with the ray's direction as inputs and predicts the ray's color.

\subsection{Grid-based Ray Representation}
\label{sec: Grid-based Ray Representation}
\begin{figure}[t!]
    \centering
    \includegraphics[width=0.8\linewidth]{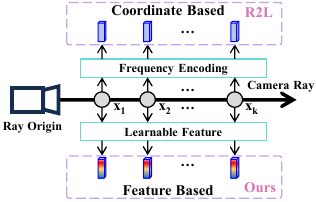}
    \vspace{-10pt}
    \caption{\textbf{Illustration of ray representation method difference} between R2L \cite{wang2022r2l} and our G-NeLF. R2L directly concatenates frequency encoded \cite{mildenhall2020nerf} coordinates as the network input. We use feature sequence to represent one ray.}
    \vspace{-13pt}
    \label{fig:Ray Representation}
\end{figure}

\vspace{-3pt}
\noindent\textbf{Ray Representation.}
Early ray representation methods~\cite{Gortler_Grzeszczuk_Szeliski_Cohen_1996, Levoy_Hanrahan_1996, sitzmann2021light, suhail2022light} primarily focus on 4-D coordinate ($(u,v), (s,t)$) representation. 
Recent approaches~\cite{wang2022r2l, cao2023real, yu2023dylin} begin to employ the $3 K$-D coordinate representation method. 
These method samples $K$ points, $\{\mathbf{x_1},...,\mathbf{x_k}\}$, along the ray to construct the representation, as illustrated in the upper part of Fig.~\ref{fig:Ray Representation}.
However, reliance solely on coordinate representation poses challenges in learning the mapping relationship between a ray's color and its coordinates, complicating the accurate prediction of color.
Consequently, these methods require extensive training data to effectively leverage the neural network's memory capacity for encoding the color of rays.

Rethinking the ignorance of utilizing the neural networks' \textit{\textbf{inference capability}}~\cite{wang2022r2l, cao2023real, yu2023dylin, gupta2024lightspeed}, we aim to develop a representation method that effectively addresses this aspect.
We introduce a $K N$-D representation, wherein \textit{\textbf{a sequence of features is employed to represent a single ray}}, as illustrated in the below of Fig.~\ref{fig:Ray Representation}.
Specifically, we orderly sample $K$ points along the ray, and extract their corresponding $N$-dimensional spatial-aware features to compose a feature sequence, $\{\mathbf{\mathcal{F}_1},...,\mathbf{\mathcal{F}_k}\}$.
The spatial awareness embedded within this feature sequence is key to enabling effective inference. 
Moreover, although there is no explicitly point-wise physical modeling, the spatial-aware feature sequence still retains the capability to constrain the view consistency.

\vspace{5pt}
\noindent\textbf{Feature Grid Design.}
To derive the feature sequence, $\{\mathbf{\mathcal{F}_1},...,\mathbf{\mathcal{F}_k}\}$, we develop a notably lightweight grid decomposition method, which only requires about 0.47 M parameters.
To guarantee that each point along the ray is endowed with distinct learnable features, we configure a conceptual grid in the space.
Each vertex of the grid holds a learnable feature vector. 
We then proceed to compute the weighted average of features for each point based on its distance to the vertices of the grid where it is situated, endowing it with a unique feature vector.
Based on our analysis, we design a hash multi-resolution tri-plane as the decomposition method for our conceptual grid.

\begin{table}[t!]
    \centering
    \footnotesize
    \tabcolsep=0.6cm
    \resizebox{0.99\linewidth}{!}{
    \begin{tabular}{l|cc} \toprule
         Mask&  Cosine Similarity $\downarrow$& PSNR $\uparrow$\\ \hline
 w/o mask& -&34.64\\
         Top-6&  0.9739& 28.08\\
         Top-10&  0.8234& 17.00\\ \hline \hline
          w/o mask& -&28.20\\
         Top-6&  0.9860& 25.74\\
         Top-9&  0.8287& 17.65\\ \bottomrule
    \end{tabular}}
    \vspace{-5pt}
    \caption{\textbf{Comparison of different numbers of feature masks}. \textit{w/o mask} means original Instant-NGP model. \textit{Top-6} means that the top 6 resolutions are masked to 0.  \textit{Top-10} means that the top 10 resolutions are masked to 0. }
    \vspace{-13pt}
    \label{tab: Hash Grid}
\end{table}

\begin{figure}[t!]
    \centering
    \includegraphics[width=1\linewidth]{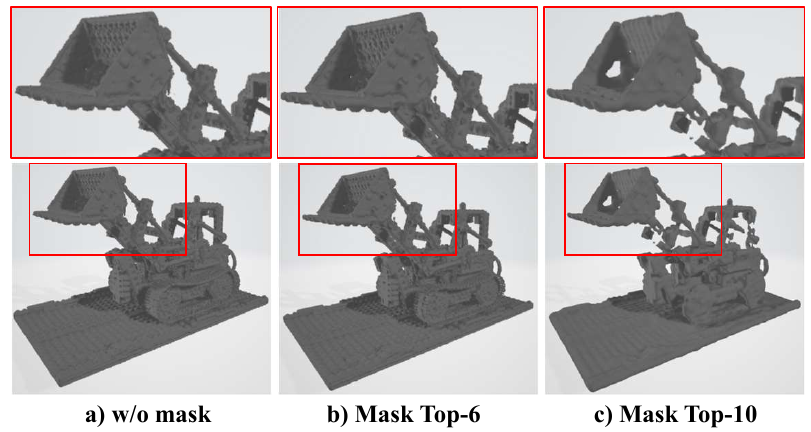}
    \vspace{-20pt}
    \caption{\textbf{Visual comparison of different numbers of feature masks}. a) Instant-NGP. b) Mask top 6 resolution in Instant-NGP. c) Mask top 10 resolution in Instant-NGP. Observing the results, it is evident that masking features at higher resolutions still allows for the preservation of overall shape information.}
    \vspace{-15pt}
    \label{fig:Hash Grid}
\end{figure}

We conduct a thorough analysis of the relationship between multi-resolution and hash mapping based on Instant-NGP~\cite{muller2022instant}, exploring the mechanisms underlying their interaction.
We argue that \textbf{\textit{the low-resolution primarily determines the shape features while the high-resolution represents the surface appearance features, self-adapting at the training stage gradually}}. 
In the hash table, the low-resolution grids are exclusively owned by themselves.
Therefore, they can reserve the structure information accurately.
On the contrary, in the hash table of the highest resolution grid, $2^{19}$ items are used to represent $2^{36}$ positions' feature, which means approximately $100,000$ positions are mapped to a single item in the hash table.
Because the positions far away from the first surface don't influence the color in the volume rendering formula~\cite{kajiya1984ray, mildenhall2020nerf}, they don't affect the corresponding color feature entries in the hash table during the training phase.
As a result, in the set of positions that are mapped to the same entry in the hash table, the position carrying the larger weight becomes the dominant one in influencing this entry following training progresses.
Consequently, the features at higher resolutions adaptively concentrate on capturing the appearance features.
To prove our argument, we conduct well-designed experiments.

To obtain the main function of different resolutions, we employ zero-masking on high-resolution features during the computation process.
As demonstrated in Fig.~\ref{fig:Hash Grid}, the shape is largely preserved even when the top six resolutions are masked.
As shown in the top half of Tab.~\ref{tab: Hash Grid}, the similarity between the original and masked hash grids is 0.9739.
However, the PSNR metric, indicative of rendered image similarity, exhibits a significant reduction.
When we mask the top ten resolutions, the shape is still relatively well-preserved with a similarity of 0.8234, but there is a more pronounced loss in color information.
A similar trend can be observed when replacing the hash grid with our hash tri-plane in the bottom part of Tab.~\ref{tab: Hash Grid}.

Though there is no explicit modeling of volume density in G-NeLF, the feature sequence is spatial-aware, which creates a huge potential to apply the same strategy.
Even, without the requirements of extra information, \ie, volume density, NeLF has a greater potential to allow a much more compact representation.
For NeLF, we design a more compact hash-based multi-resolution tri-plane representation that yields superior results compared with the hash grid.
As depicted in Fig.~\ref{fig:Framework}, our process begins by projecting the points from the ray onto three orthogonal planes. 
Subsequently, according to the projected coordinates, we identify the corresponding grid vertices across all resolutions for each feature plane.
Then, we get the vertices' features of each resolution from a hash table.
The hash table mapping function is

\vspace{-10pt}
\begin{equation}
    Index = (x a~\oplus~y b) \mod T
\vspace{-4pt}
\end{equation}

where $x$ and $y$ are the indexes of height and width, $a=1$, $b=2,654,435,761$, $\oplus$ is the XOR opration, $T$ is the hash table size.
We use the $Index$ to retrieve the specific feature vector from the corresponding hash table.
Subsequently, we calculate the weighted-sum feature based on the distance between the point and these vertices.
Finally, we concatenate the feature vectors from all resolutions as the feature of that point.
The concrete procedure of calculation is illustrated in Algorithm \ref{alg: point feature}.

\vspace{-3pt}
\subsection{Ray Color Decoder}
\label{sec: Ray Color Decoder}
\vspace{-3pt}
Once we acquire the feature sequence, $\{\mathbf{\mathcal{F}_1},...,\mathbf{\mathcal{F}_k}\}$, we can predict its color from this sequence.
A direct approach is to employ an MLP network to directly predict the color.
However, due to the neglect of the potent temporal-order sequential information remaining in the feature sequence, the MLP network is not efficient in predicting the color.
We also attempt to employ a positional-aware Transformer-based~\cite{vaswani2017attention} architecture.
Nevertheless, as same as the analysis in~\cite{zeng2023transformers}, since the calculation of attention is permutation-invariant and anti-order, it is not suitable for our target either.
Despite there being some position encoding techniques, there remains an inevitable loss of certain temporal-order sequential features.

\begin{figure}[t!]
\vspace{-10pt}
\begin{algorithm}[H]
	\caption{Point's Feature Acquirement}
	\label{alg: point feature} 
	\begin{algorithmic}[1]
	    \STATE \textbf{Input}: $x$, the 3D coordinate of input point; $\{\pi_{xy}$, $\pi_{xz}$, $\pi_{yz} \}$, the tri-plane; $s_i$, the grid size of $i$-th resolution; $N$, the number of resolutions of each plane; $hash_{\pi,i}(\cdot)$, the hash table of $i$-th resolution in the plane $\pi$;
        \STATE \textbf{Output}: $F$, feature vector of the input point $x$; \vspace{2pt}
	    \FOR{$\pi$ in $\{\pi_{xy}, \pi_{xz}, \pi_{yz} \}$} \vspace{3pt}
            \STATE $F_{\pi}$ = []; \vspace{4pt} ~~\textcolor{gray}{\# empty list}
            \FOR{$i$ $\xleftarrow[]{}$ 1 to $N$} \vspace{4pt}
                \STATE  $x_{\pi}$ = Projection(x, $\pi$);  ~~\textcolor{gray}{\# 2D Coordinate} \vspace{3pt}
                \STATE  $t\_l$ = $hash_{\pi,i}$([$x_{\pi}[0]$ // $s_i$, $x_{\pi}[1]$ // $s_i$]); \vspace{3pt}
                \STATE  $t\_r$ = $hash_{\pi,i}$([$x_{\pi}[0]$ // $s_i$, $x_{\pi}[1]$ // $s_i$ + 1]); \vspace{3pt}
                \STATE  $b\_l$ = $hash_{\pi,i}$([$x_{\pi}[0]$ // $s_i$ + 1, $x_{\pi}[1]$ // $s_i$]); \vspace{3pt}
                \STATE  $b\_r$ = $hash_{\pi,i}$([$x_{\pi}[0]$ // $s_i$ + 1, $x_{\pi}[1]$ // $s_i$ + 1]); \vspace{3pt}
    	      \STATE  $F_{\pi}$.append(Weighted\_Sum([$t\_l$, $t\_r$, $b\_l$, $b\_r$])); \vspace{3pt}
            \ENDFOR \vspace{3pt} 
            \STATE $F_{\pi}$ = Concat($F_{\pi}$); \vspace{3pt}
    	 \ENDFOR \vspace{3pt}
        \STATE $F$ = Concat([$F_{\pi_{xy}}$, $F_{\pi_{xz}}$, $F_{\pi_{yz}}$]); \vspace{3pt}
	    \STATE  \textbf{return}  $F$; \vspace{3pt}
	    \STATE  \textbf{End}.
	\end{algorithmic} 

\end{algorithm}
\vspace{-25pt}
\end{figure}

Considering that ray propagation in the real world follows a path from near to far, we propose employing a lightweight LSTM network~\cite{hochreiter1997long} to emulate this sequential process, as shown in Fig.~\ref{fig:Framework} (right).
Unlike text sequences, the temporal order of a ray's propagation is crucial.
The results presented in Tab.~\ref{tab:Inference Module Ablation} demonstrate that recurrent computation is a critical component enabling the effectiveness of our ray color decoder.
Andrej~\cite{karpathy2015visualizing} \etal argue that LSTM with more than two layers exhibits fundamental differences from its single-layer counterparts.
Consistent with this assertion, our experimental results (Fig.~\ref{fig:LSTM Experiments}(a)) indicate that a 2-layer LSTM surpasses the performance of a 1-layer LSTM when computational costs are equivalent.
To simulate the view-dependent phenomenon, we additionally incorporate the ray direction into the decoder.
We employ Sphere Harmonics, as adopted by Ref-NeRF~\cite{verbin2022ref} and Instant-NGP~\cite{muller2022instant}, to encode the 3D normalized direction vector.
The encoded direction is concatenated to the ray sequential feature.
Formally, the decoding process can be written as 
\begin{equation}
    \mathcal{H}_{i}, \mathcal{C}_{i} = LSTM(Concat(\mathcal{F}_{i}, d), \mathcal{H}_{i-1}, \mathcal{C}_{i-1}),
\end{equation}
\begin{equation}
    (r, g, b) = MLP(\mathcal{H}_{k}),
\end{equation}
where $i=1,2,...,k$, $\mathcal{F}_i$ is coming from ray's feature sequence $\{\mathbf{\mathcal{F}_1},...,\mathbf{\mathcal{F}_k}\}$, $d$ is the encoded ray's direction, $\mathcal{H}_{i}$ and $\mathcal{C}_{i}$ are hidden state and cell state in LSTM, $LSTM(\cdot)$ is a LSTM network, $Concat(\cdot)$ is the concatenation function, $MLP(\cdot)$ is a two-layer MLP network, $(r,g,b)$ is the target, ray's color.

\vspace{-5pt}
\section{Experiments}

\vspace{-3pt}
\subsection{Datasets and Implementation Details}
\vspace{-3pt}

\noindent\textbf{Datasets.}
We use two datasets to evaluate, including Realistic Synthetic 360° and Real Forward-Facing~\cite{mildenhall2020nerf}.
The Realistic Synthetic 360° comprises 8 scenes.
Each scene provides 100 training views, 100 validation views, and 200 test views, all of which have a resolution of $800\times800$.
To compare with R2L~\cite{wang2022r2l}, we downsample all images by 2x during training and testing.
To further evaluate our method at high resolution, we use the original resolution to compare with NeRF methods.
The Real Forward-Facing also consists of 8 scenes, with each scene containing between 20 to 62 images.
As same as other methods, we hold every 8th image for the test.
We downsample all images in Forward-Facing by a factor of 8.

\noindent\textbf{Implementation Details.} All our experiments were conducted on a single RTX3090 GPU using float16 accuracy.
We provide three configurations for experiments of 360° scenes.
For the small and medium model settings, we use a grid with 8 levels' resolution from 16 to 1024, the max hash table size is $2^{14}$, and the feature dimension of each level is 2. 
The ray color decoder is a two-layer LSTM network with 32 (small) and 128 (medium) hidden sizes.
For the large version, we use 16 levels' resolution with 16 to 2048, and the max hash table size is $2^{16}$. 
The ray color decoder has three layers with 128 hidden sizes. 
For 360° scenes, the feature sequence length is set at 256.
For forward-facing scenes, we convert the coordinate system to NDC and establish the feature sequence length at 128.
We use the MSE color loss to supervise the whole model in an end-to-end manner.


\begin{table*}[t!]

 \vspace{-5pt}
    \centering
    \tabcolsep=0.3cm
    \resizebox{\linewidth}{!}{
\begin{tabular}{l|c|c|c|c|c|c|c|c|c|cc}
\toprule
Method & Avg & Chair & Drums & Ficus & Hotdog & Lego & Materials & Mic & Ship & Size(MB) & Data Size \\ \hline
NeRF~\cite{mildenhall2020nerf} & 30.47 & 33.90 & 25.56 & 28.88 & 34.64 & 31.42 & 29.22 & 30.84 & 29.30 & 4.56 & 100 \\
R2L~\cite{wang2022r2l} & 31.87 & \cellcolor[HTML]{F4B5B4}36.71 & \cellcolor[HTML]{F4B5B4}26.03 & 28.63 & \cellcolor[HTML]{FFFFBB}38.07 & 32.53 & 30.20 & 32.80 & \cellcolor[HTML]{FFFFBB}29.98 & 22.62 & 10,100 \\
Ours-S & \cellcolor[HTML]{FFFFBB}31.91 & 34.39 & 24.81 & \cellcolor[HTML]{FFFFBB}29.40 & 37.62 & \cellcolor[HTML]{FFFFBB}33.83 & \cellcolor[HTML]{FFFFBB}30.54 & \cellcolor[HTML]{FFFFBB}34.78 & 29.93 & 0.95 & 100 \\
Ours-M & \cellcolor[HTML]{F9DAB7}33.89 & \cellcolor[HTML]{F9DAB7}36.65 & \cellcolor[HTML]{FFFFBB}25.86 & \cellcolor[HTML]{F9DAB7}30.82 & \cellcolor[HTML]{F9DAB7}39.40 & \cellcolor[HTML]{F9DAB7}36.53 & \cellcolor[HTML]{F9DAB7}33.54 & \cellcolor[HTML]{F9DAB7}37.05 & \cellcolor[HTML]{F9DAB7}31.30 & 1.41 & 100 \\
Ours-L & \cellcolor[HTML]{F4B5B4}34.35 & \cellcolor[HTML]{FFFFBB}36.64 & \cellcolor[HTML]{F9DAB7}25.91 & \cellcolor[HTML]{F4B5B4}32.13 & \cellcolor[HTML]{F4B5B4}39.92 & \cellcolor[HTML]{F4B5B4}37.56 & \cellcolor[HTML]{F4B5B4}33.55 & \cellcolor[HTML]{F4B5B4}37.08 & \cellcolor[HTML]{F4B5B4}32.01 & 7.16 & 100 \\ \bottomrule
\end{tabular}}
\vspace{-10pt}
\caption{PSNR quantitative comparison at $\mathbf{400\times400}$ resolution on Realistic Synthetic 360° dataset. \textit{Size} is model size. \textit{Data Size} is the number of the training images.  \colorbox[HTML]{F4B5B4}{value} means first, \colorbox[HTML]{F9DAB7}{value} means second, \colorbox[HTML]{FFFFBB}{value} means third.}

    \label{tab:Comparison with R2L on 360}
\end{table*}

\begin{table*}[t!]

 \vspace{-5pt}
\footnotesize
\centering
\tabcolsep=0.2cm
\resizebox{\linewidth}{!}{
\begin{tabular}{c|l|c|c|c|c|c|c|c|c|c|cc}
\toprule
Category&Method & Avg & Chair & Drums & Ficus & Hotdog & Lego & Materials & Mic & Ship & Size(MB) & Data Size \\ \hline
\multirow{2}{*}{Compression}&VQRF~\cite{li2023compressing} & 31.77 & 33.80 & 25.38 & 32.67 & 36.47 & 34.27 & 29.28 & 33.11 & 29.24 & $\overline{1.43}$ & 100 \\ 
&\textbf{Ours-M} & 32.56 & 33.41 & 24.91 & 31.78 & \cellcolor[HTML]{F9DAB7}37.72 & 35.41 & \cellcolor[HTML]{F9DAB7}32.25 & 35.73 & 29.24 & \textbf{1.41} & 100 \\ \hline \hline
\multirow{4}{*}{NeRF}&NeRF~\cite{mildenhall2020nerf} & 31.01 & 33.00 & 25.01 & 30.13 & 36.18 & 32.54 & 29.62 & 32.91 & 28.65 & 5 & 100 \\
&Mip-NeRF~\cite{barron2021mip} & 33.09 & \cellcolor[HTML]{F9DAB7}35.14 & 25.48 & \cellcolor[HTML]{FFFFBB}33.29 & \cellcolor[HTML]{FFFFBB}37.48 & 35.70 & \cellcolor[HTML]{FFFFBB}30.71 & \cellcolor[HTML]{F4B5B4}36.51 & \cellcolor[HTML]{FFFFBB}30.41 & 2.5 & 100 \\
&TensoRF~\cite{chen2022tensorf} & \cellcolor[HTML]{FFFFBB}33.14 & \cellcolor[HTML]{F4B5B4}35.76 & \cellcolor[HTML]{F9DAB7}26.01 & \cellcolor[HTML]{F4B5B4}33.99 & 37.41 & \cellcolor[HTML]{F9DAB7}36.46 & 30.12 & 34.61 & \cellcolor[HTML]{F9DAB7}30.77 & 71.8 & 100 \\
&I-NGP~\cite{muller2022instant} & \cellcolor[HTML]{F9DAB7}33.18 & \cellcolor[HTML]{FFFFBB}35.00 & \cellcolor[HTML]{F4B5B4}26.02 & \cellcolor[HTML]{F9DAB7}33.51 & 37.40 & \cellcolor[HTML]{FFFFBB}36.39 & 29.78 & \cellcolor[HTML]{FFFFBB}36.22 & \cellcolor[HTML]{F4B5B4}31.10 & 64.1 & 100 \\ \hline \hline
\multirow{3}{*}{NeLF}&MobileR2L~\cite{cao2023real} & 31.34 & 33.66 & 25.05 & 29.80 & 36.84 & 32.18 & 30.54 & 34.37 & 28.75 & 8.20 & 10,100 \\
&LightSpeed~\cite{gupta2024lightspeed} & 32.23 & 34.21 & \cellcolor[HTML]{FFFFBB}25.63 & 32.82 & 36.77 & 34.35 & 29.51 & 35.65 & 28.90 & 13.60 & 30,100 \\
&\textbf{Ours-L} & \cellcolor[HTML]{F4B5B4}33.19 & 34.56 & 25.01 & 32.45 & \cellcolor[HTML]{F4B5B4}38.03 & \cellcolor[HTML]{F4B5B4}36.78 & \cellcolor[HTML]{F4B5B4}32.32 & \cellcolor[HTML]{F9DAB7}36.23 & 30.10 & 7.16 & 100 \\ \bottomrule
\end{tabular}}
\vspace{-10pt}
\caption{PSNR quantitative comparison at $\mathbf{800\times800}$ resolution on Realistic Synthetic 360° dataset. The VQRF is based on DVGO~\cite{sun2022direct}. \textit{Size} is model size. \textit{Data Size} is the number of training images. \textbf{The first part} is NeRF's compression method. \textbf{The second part} is the SoTA NeRF methods. \textbf{The third part} is the SoTA NeLF methods.}

    \label{tab:Comparison with Other Methods on 360}
\end{table*}

\begin{table*}[t!]

 \vspace{-5pt}
    \centering
    \tabcolsep=0.15cm
    \resizebox{\linewidth}{!}{
    \begin{tabular}{l|cccccc|ccc|c}
\toprule
\multirow{2}{*}{Method} & \multicolumn{6}{c|}{Dense} & \multicolumn{3}{c|}{Sparse} & \multirow{2}{*}{\begin{tabular}[c]{@{}c@{}}Additional\\ Images\end{tabular}} \\ \cline{2-10}
 & \multicolumn{1}{c|}{Avg} & \multicolumn{1}{c|}{Horns(54)} & \multicolumn{1}{c|}{Trex(48)} & \multicolumn{1}{c|}{Room(36)} & \multicolumn{1}{c|}{Fortress(36)} & Flower(29) & \multicolumn{1}{c|}{Leaves(21)} & \multicolumn{1}{c|}{Orchids(21)} & Fern(17) &  \\ \hline
NeRF~\cite{mildenhall2020nerf} & \multicolumn{1}{c|}{30.17} & \multicolumn{1}{c|}{\cellcolor[HTML]{FFFFBB}28.86} & \multicolumn{1}{c|}{28.10} & \multicolumn{1}{c|}{33.07} & \multicolumn{1}{c|}{\cellcolor[HTML]{F9DAB7}32.61} & 28.22 & \multicolumn{1}{c|}{\cellcolor[HTML]{F9DAB7}22.40} & \multicolumn{1}{c|}{\cellcolor[HTML]{F9DAB7}21.29} & \cellcolor[HTML]{F4B5B4}26.86 & 0 \\
Attal \etal~\cite{attal2022learning} & \multicolumn{1}{c|}{\cellcolor[HTML]{F9DAB7}30.65} & \multicolumn{1}{c|}{\cellcolor[HTML]{F9DAB7}30.12} & \multicolumn{1}{c|}{\cellcolor[HTML]{F4B5B4}29.41} & \multicolumn{1}{c|}{\cellcolor[HTML]{F4B5B4}33.57} & \multicolumn{1}{c|}{31.46} & \cellcolor[HTML]{F9DAB7}28.71 & \multicolumn{1}{c|}{21.82} & \multicolumn{1}{c|}{\cellcolor[HTML]{FFFFBB}20.29} & 24.25 & 0 \\
R2L~\cite{wang2022r2l} & \multicolumn{1}{c|}{\cellcolor[HTML]{FFFFBB}30.35} & \multicolumn{1}{c|}{28.95} & \multicolumn{1}{c|}{\cellcolor[HTML]{FFFFBB}28.12} & \multicolumn{1}{c|}{\cellcolor[HTML]{F9DAB7}33.30} & \multicolumn{1}{c|}{\cellcolor[HTML]{F4B5B4}32.71} & \cellcolor[HTML]{FFFFBB}28.67 & \multicolumn{1}{c|}{\cellcolor[HTML]{F4B5B4}22.71} & \multicolumn{1}{c|}{\cellcolor[HTML]{F4B5B4}21.46} & \cellcolor[HTML]{F9DAB7}26.87 & 10,000 \\
Ours & \multicolumn{1}{c|}{\cellcolor[HTML]{F4B5B4}30.97} & \multicolumn{1}{c|}{\cellcolor[HTML]{F4B5B4}30.55} & \multicolumn{1}{c|}{\cellcolor[HTML]{F9DAB7}29.30} & \multicolumn{1}{c|}{\cellcolor[HTML]{FFFFBB}33.28} & \multicolumn{1}{c|}{\cellcolor[HTML]{FFFFBB}32.46} & \cellcolor[HTML]{F4B5B4}29.27 & \multicolumn{1}{c|}{\cellcolor[HTML]{FFFFBB}22.04} & \multicolumn{1}{c|}{19.46} & \cellcolor[HTML]{FFFFBB}25.29 & 0 \\ \bottomrule
\end{tabular}}
\vspace{-10pt}
 \caption{PSNR quantitative comparison at $\mathbf{504\times378}$ resolution on Real Forward-Facing. \textit{Additional Images} is the required number of extra synthesized images. The number of original training views is in the brackets.}

    \label{tab:Comparison with R2L on Forward-Facing}
\end{table*}

\begin{table*}[t!]
    \centering
    \tabcolsep=0.1cm

     \vspace{-5pt}
    \resizebox{\linewidth}{!}{
    \begin{tabular}{c|l|cccccc|ccc|c}
\toprule
\multirow{2}{*}{Category}&\multirow{2}{*}{Method} & \multicolumn{6}{c|}{Dense} & \multicolumn{3}{c|}{Sparse} &  \\ \cline{3-11}
 && \multicolumn{1}{c|}{Avg} & \multicolumn{1}{c|}{Horns(54)} & \multicolumn{1}{c|}{Trex(48)} & \multicolumn{1}{c|}{Room(36)} & \multicolumn{1}{c|}{Fortress(36)} & Flower(29) & \multicolumn{1}{c|}{Leaves(21)} & \multicolumn{1}{c|}{Orchids(21)} & Fern(17) & \multirow{-2}{*}{\begin{tabular}[c]{@{}c@{}}Additional\\ Images\end{tabular}} \\ \hline \hline
\multirow{3}{*}{NeRF}&NeRF~\cite{mildenhall2020nerf} & \multicolumn{1}{c|}{\cellcolor[HTML]{FFFFBB}29.10} & \multicolumn{1}{c|}{27.45} & \multicolumn{1}{c|}{\cellcolor[HTML]{FFFFBB}26.80} & \multicolumn{1}{c|}{\cellcolor[HTML]{F4B5B4}32.70} & \multicolumn{1}{c|}{\cellcolor[HTML]{F9DAB7}31.16} & 27.40 & \multicolumn{1}{c|}{20.92} & \multicolumn{1}{c|}{\cellcolor[HTML]{F4B5B4}20.36} & \cellcolor[HTML]{FFFFBB}25.17 & 0 \\
&Plenoxels~\cite{fridovich2022plenoxels} & \multicolumn{1}{c|}{28.64} & \multicolumn{1}{c|}{\cellcolor[HTML]{FFFFBB}27.58} & \multicolumn{1}{c|}{26.48} & \multicolumn{1}{c|}{30.22} & \multicolumn{1}{c|}{31.09} & 27.83 & \multicolumn{1}{c|}{\cellcolor[HTML]{F4B5B4}21.41} & \multicolumn{1}{c|}{\cellcolor[HTML]{FFFFBB}20.24} & \cellcolor[HTML]{F4B5B4}25.46 & 0 \\
&TensoRF~\cite{chen2022tensorf} & \multicolumn{1}{c|}{29.08} & \multicolumn{1}{c|}{\cellcolor[HTML]{F9DAB7}27.64} & \multicolumn{1}{c|}{26.61} & \multicolumn{1}{c|}{31.80} & \multicolumn{1}{c|}{\cellcolor[HTML]{FFFFBB}31.14} & \cellcolor[HTML]{F9DAB7}28.22 & \multicolumn{1}{c|}{\cellcolor[HTML]{F9DAB7}21.34} & \multicolumn{1}{c|}{20.02} & \cellcolor[HTML]{F9DAB7}25.31 & 0 \\ \hline \hline
\multirow{3}{*}{NeLF}&MobileR2L~\cite{cao2023real} & \multicolumn{1}{l|}{28.85} & \multicolumn{1}{c|}{27.01} & \multicolumn{1}{c|}{26.71} & \multicolumn{1}{c|}{\cellcolor[HTML]{FFFFBB}32.09} & \multicolumn{1}{c|}{30.81} & 27.61 & \multicolumn{1}{c|}{20.52} & \multicolumn{1}{c|}{20.06} & 24.39 & 10,000 \\
&LightSpeed~\cite{gupta2024lightspeed} & \multicolumn{1}{l|}{\cellcolor[HTML]{F9DAB7}29.12} & \multicolumn{1}{c|}{27.04} & \multicolumn{1}{c|}{\cellcolor[HTML]{F9DAB7}26.93} & \multicolumn{1}{c|}{\cellcolor[HTML]{F9DAB7}32.32} & \multicolumn{1}{c|}{\cellcolor[HTML]{F4B5B4}31.45} & \cellcolor[HTML]{FFFFBB}27.88 & \multicolumn{1}{c|}{\cellcolor[HTML]{FFFFBB}21.01} & \multicolumn{1}{c|}{\cellcolor[HTML]{F9DAB7}20.33} & 25.05 & 10,000 \\
&Ours & \multicolumn{1}{c|}{\cellcolor[HTML]{F4B5B4}29.26} & \multicolumn{1}{c|}{\cellcolor[HTML]{F4B5B4}28.08} & \multicolumn{1}{c|}{\cellcolor[HTML]{F4B5B4}27.20} & \multicolumn{1}{c|}{31.89} & \multicolumn{1}{c|}{30.77} & \cellcolor[HTML]{F4B5B4}28.35 & \multicolumn{1}{c|}{20.88} & \multicolumn{1}{c|}{19.01} & 23.68 & 0 \\ \bottomrule
\end{tabular}}
\vspace{-10pt}
\caption{PSNR quantitative comparison at $\mathbf{1008\times756}$ resolution on Real Forward-Facing. \textit{Additional Images} is the required number of extra synthesized images. The number of original training views is in the brackets.}
\vspace{-3pt}
\label{tab:Comparison with R2L on Forward-Facing 1008}
\end{table*}

\vspace{-5pt}
\subsection{Overall Comparison}
\begin{figure}[t!]
    \centering
    \vspace{-5pt}
    \includegraphics[width=1.0\linewidth]{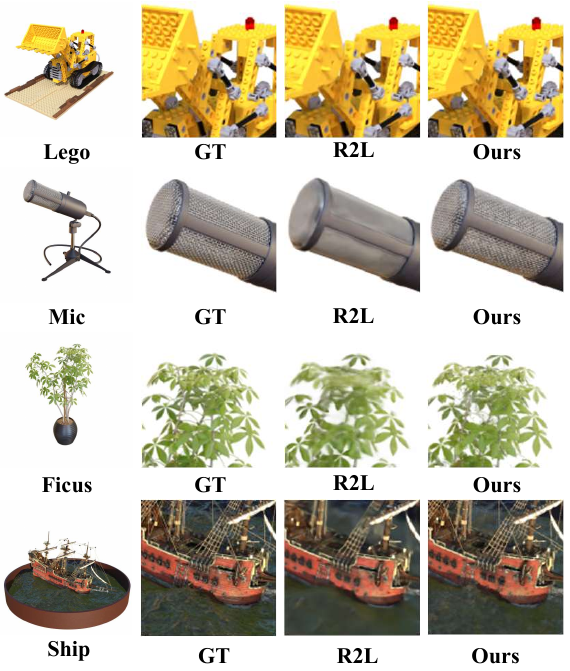}
    \vspace{-20pt}
    \caption{\textbf{Visual comparison} with R2L~\cite{wang2022r2l} on \textit{Lego}, \textit{Mic}, \textit{Ficus}, and \textit{Ship} scene in Realistic Synthetic 360° dataset.}
    \vspace{-12pt}
    \label{fig:Blender}
\end{figure}

\vspace{-3pt}
\noindent\textbf{Realistic Synthetic 360° Dataset.} 
We first compare our method with the NeLF method, R2L~\cite{wang2022r2l}, at $400 \times 400$.
In Tab.~\ref{tab:Comparison with R2L on 360}, we show PSNR and SSIM comparison results scene by scene.
Different from the training difficulty and resource consumption that R2L needs, we only need the original training data, such as 100 views, saving over 56 GB space.
This advantage also enables our method to train models on higher-resolution data effectively.
Despite our method's minimal resource consumption, it significantly outperforms R2L, underscoring its superiority.
Our small configuration, utilizing just 100 views for training, still surpasses R2L 0.04 dB with only a 0.95 MB model size.
Furthermore, continuing to scale up the model size results in a consistent improvement, whereas the performance of R2L has reached its saturation point.
In Fig.~\ref{fig:Blender}, we present a visual comparative analysis between R2L and Ours.
For example, in the \textit{Lego} scene, R2L encounters difficulties rendering the region behind the shovel, producing an image with compromised clarity. 
Our method, conversely, renders this area with distinct sharpness, showcasing its enhanced performance.

As our G-NeLF is a lightweight model, we also compare it with the SoTA NeRF model compression method, VQRF \cite{li2023compressing}.
As Tab.~\ref{tab:Comparison with Other Methods on 360} shown, when evaluated under comparable model size, 1.43 MB for \textit{DVGO-VQDF} and 1.41 MB for \textit{Ours-M}, our method consistently delivers superior results by \textbf{+0.79} PSNR. 
This large gain shows the efficiency of our tiny hash tri-plane design and ray color decoder.

To demonstrate that we have pushed the NeLF far beyond the performance of previous methods, we also compare it with some SoTA grid-based NeRF methods, including TensoRF~\cite{chen2022tensorf} and Instant-NGP~\cite{muller2022instant}.
As shown in Tab.~\ref{tab:Comparison with Other Methods on 360}, our method can achieve comparable results on par with SoTA NeRF techniques, which proves that NeLF also has a large potential on novel view synthesis tasks with relatively low resources.
Grid-based NeRFs usually introduce a lot of storage consumption of model size, \eg, 64.1MB of Instant-NGP~\cite{muller2022instant} and 71.8MB of TensoRF~\cite{chen2022tensorf}.
Our ray-wise representation, which does not rely on precise physical modeling, enables us to represent a scene with a much more compact structure, about one-tenth parameter usage.

\begin{figure}[t!]
    \centering
    \includegraphics[width=1.0\linewidth]{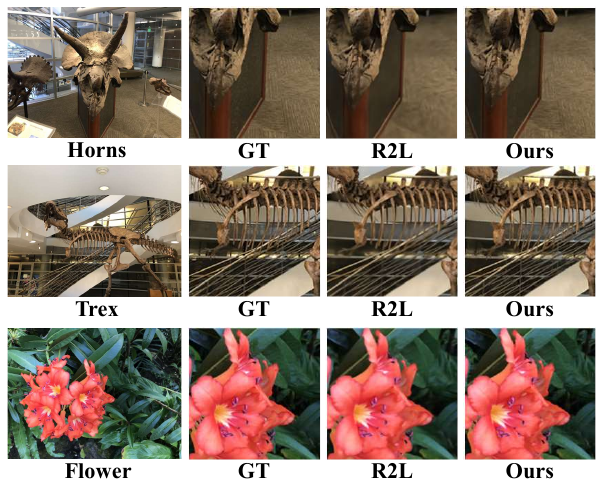}
    \vspace{-20pt}
    \caption{\textbf{Visual comparison} with R2L~\cite{wang2022r2l} on \textit{Horns}, \textit{Trex}, and \textit{Flower} scene in Real Forward-Facing dataset.}
    \vspace{-8pt}
    \label{fig:LLFF}
\end{figure}

\noindent\textbf{Real Forward-Facing Dataset.}
As shown in Tab.~\ref{tab:Comparison with R2L on Forward-Facing}, we first compare our method with NeRF~\cite{mildenhall2020nerf}, NeLF~\cite{attal2022learning}, and R2L~\cite{wang2022r2l} on the Real Forward-Facing dataset at $\mathbf{504\times378}$ resolution. 
At the head of the table, the number of original training views is in the brackets.
Moreover, the results at $\mathbf{1008\times756}$ resolution are shown in Tab.~\ref{tab:Comparison with R2L on Forward-Facing 1008}, including NeRF-based methods~\cite{li2023compressing, mildenhall2020nerf, fridovich2022plenoxels, chen2022tensorf} and NeLF-based methods~\cite{cao2023real, gupta2024lightspeed}.
Unlike the meticulously constructed Realistic Synthetic 360° dataset, the Real Forward-Facing dataset, sourced from the real world, has inherent inaccuracies in camera parameters.
This dataset also presents a higher scene complexity. 
We attribute the superiority over other methods to the fact that our technique does not hinge on point-wise modeling, allowing for some leniency in camera parameter estimation. 
Our method is fundamentally rooted in ray-level representations, which reduces the emphasis on exact camera parameter accuracy. 
In Fig.~\ref{fig:LLFF}, we show the visual comparison with R2L.

\vspace{-5pt}
\subsection{Ablation Studies}
\vspace{-5pt}

\noindent\textbf{Design of Feature Grid.}
\label{sec: Grid Design}
To prove the superiority of our hash tri-plane design for NeLF, we replace it with the grid designs in Instant-NGP \cite{muller2022instant} and TensoRF \cite{chen2022tensorf} for comparison.
All experiments are based on our $3$-layer $128$-width ray color decoder, which ensures ample capacity to ensure that the decoder does not become a bottleneck.
The feature sequence length is set as $256$.
As Tab.~\ref{tab:Grid Ablation} shows, our method still outperforms the two methods with such a small number of parameters.
We believe this is due to more compact representations and easier training with fewer parameters.

\begin{table}[t!]
\footnotesize
    \centering
    \tabcolsep=0.6cm
    \resizebox{0.99\linewidth}{!}{
    \begin{tabular}{c|cc} \hline
         Method&  PSNR&  Params\\ \hline
         Instant-NGP&  34.83&  12,239,728\\
         TensoRF&  34.85&  13,003,200\\
         Ours&  \textbf{35.36}&  \textbf{474,432}\\ \hline
    \end{tabular}}
    \vspace{-8pt}
    \caption{\textbf{The grid design comparison} among Instant-NGP-based \cite{muller2022instant}, TensoRF-based \cite{chen2022tensorf}, and our hash tri-plane for our G-NeLF.}
    \vspace{-8pt}
    \label{tab:Grid Ablation}
\end{table}

\noindent\textbf{Design of Ray Color Decoder.}
In Tab.~\ref{tab:Inference Module Ablation}, we present a comparison among different designs of ray color decoders, ensuring a similar computational load across all methods for fairness.
Our experiments span many prevalent network architectures, including MLP, RNN, GRU~\cite{Cho_vanMerrienboer_Gulcehre_Bahdanau_Bougares_Schwenk_Bengio_2014}, Transformer~\cite{vaswani2017attention}, and LSTM~\cite{hochreiter1997long}. 
Our findings reveal that GRU and LSTM-based ray color decoders are particularly effective, validating our hypothesis that temporal sequential information is the key factor in inferring the ray feature sequence effectively. 
Moreover, although Transformer-based decoders incorporate position encoding techniques to perceive temporal relationships, the performance is still unsatisfactory for a lot of temporal prediction tasks~\cite{zeng2023transformers}.

\begin{table}[t!]
\footnotesize
    \centering
    \tabcolsep=0.3cm
    \resizebox{0.99\linewidth}{!}{
    \begin{tabular}{c|ccc} \hline
         Method&  PSNR&  Params &FPS\\ \hline
         MLP-Based&  22.35&  71,571,459&4.55\\
 RNN-Based& 15.52& 25,347&6.54\\
         Transformer-Based&  31.56&   21,603&1.32\\
 GRU-Based& 33.07& 18,051&5.71\\
         Ours-LSTM Design&  \textbf{33.82}&  23,299&\textbf{6.71}\\ \hline
    \end{tabular}}
    \vspace{-8pt}
    \caption{\textbf{The validation} of our ray color decoder design.}
    \vspace{-8pt}
    \label{tab:Inference Module Ablation}
\end{table}

\begin{table}[t!]
\footnotesize
    \centering
    \tabcolsep=0.8cm
    \resizebox{0.99\linewidth}{!}{
    \begin{tabular}{c|c} \hline
         Method&  PSNR\\ \hline
         w/o direction&  33.45\\
         Frequency Encoding&  33.40(-0.05)\\
 Spherical Harmanics& 33.82(+0.37)\\ \hline
    \end{tabular}}
    \vspace{-8pt}
    \caption{\textbf{The different encoding methods} for direction vector.}
    \vspace{-13pt}
    \label{tab:Dir Encoder}
\end{table}

\noindent\textbf{The Effect of Ray Direction.}
To ascertain the influence of ray direction on the performance of the ray color decoder, we conduct the ablation experiments under our G-NeLF small model.
We compare the different direction vector encoding methods, including Frequency Encoding used in vanilla NeRF \cite{mildenhall2020nerf} and Spherical Harmonics used in Instant-NGP \cite{muller2022instant}.
As shown in Tab.~\ref{tab:Dir Encoder}, Spherical Harmonics is more suitable for our G-NeLF.

\begin{figure}[t!]
    \centering
    \includegraphics[width=1.0\linewidth]{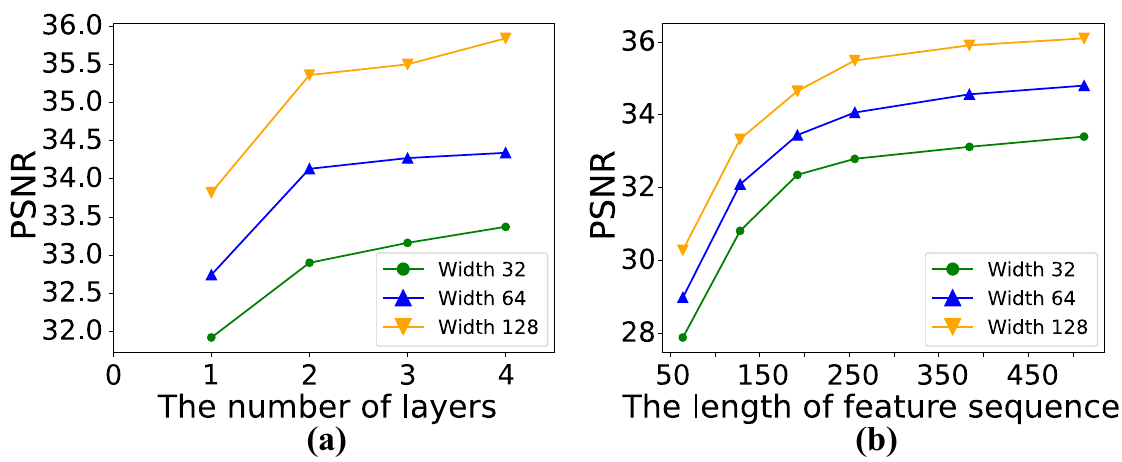}
    \vspace{-20pt}
    \caption{(a) The ablation study of the number of layers. (b) The ablation study of the length of feature sequence.}
    \vspace{-13pt}
    \label{fig:LSTM Experiments}
\end{figure}

\noindent\textbf{The Number of LSTM Layers.}
According to the research of Andrej \etal~\cite{karpathy2015visualizing}, the LSTM network with more than two layers exhibits fundamental differences and significant performance enhancements compared with its single-layer counterparts. 
We use the small configuration grid setting and we set the sequence length as $256$.
Fig.~\ref{fig:LSTM Experiments} (a) presents a comparative analysis of the impact of varying the number of layers across three different width settings: $32$, $64$, and $128$.
We can observe that there is an obvious gap between $1$-layer and $2$-layer under all three width settings.
However, if we deepen the number of layers, the gain is not remarkable.
In light of these findings and taking into account the linear increase in computational overhead associated with additional layers, we apply two layers for both our small and medium configurations. 

\noindent\textbf{The Length of Ray Feature Sequence.}
We proceed to investigate the impact of feature sequence length across three distinct settings of LSTM width. The results of this exploration are systematically presented in Fig.~\ref{fig:LSTM Experiments} (b).
A discernible trend emerges from the data, indicating that as the length of the feature sequence increases, there is a corresponding and gradual enhancement in image quality.
To balance the computation afford and quality, we set $256$ as the sequence length for Realistic Synthetic 360°.

\vspace{-5pt}
\section{Conclusion}
\vspace{-5pt}

In this paper, we proposed G-NeLF, a completely implicit NeLF method for novel view synthesis. G-NeLF does not rely on any physical modeling techniques such as volume density. 
Our method demonstrated that algorithms based on volume rendering, \eg, NeRF, aren't the sole avenues to address this challenge. 
Extensive experiments demonstrate that our method, even with a modest model size of 0.95 MB and utilizing only 100 views, still can successfully surpass the SoTA NeLF, R2L~\cite{wang2022r2l}, training within approximately one hour. 
We hope that the comprehensive enhancements introduced in our method can contribute to the advancement of research in this area. 

\noindent \textbf{Future Work:} 
It remains a promising research direction to extend our method to cater to dynamic scenes.

{
    \small
    \bibliographystyle{ieeenat_fullname}
    \bibliography{main}
}
\clearpage
\setcounter{page}{1}
\maketitlesupplementary


\section{More comparison results}

We show SSIM comparison results in the following tables. Due to the SSIM calculation bug in MobileR2L~\cite{cao2023real} and LightSpeed~\cite{gupta2024lightspeed}, we don't provide their SSIM. Please refer to the PSNR report for comparison.

\begin{table*}

    \centering
    \tabcolsep=0.2cm
    \resizebox{\linewidth}{!}{
\begin{tabular}{l|c|c|c|c|c|c|c|c|c|cc}
\toprule
Method & AVG & Chair & Drums & Ficus & Hotdog & Lego & Materials & Mic & Ship & Size(MB) & Data Size \\ \hline
NeRF~\cite{mildenhall2020nerf} & 0.964 & 0.976 & 0.945 & 0.968 & 0.983 & 0.973 & 0.967 & 0.974 & 0.926 & 4.56 & 100 \\
R2L~\cite{wang2022r2l} & \cellcolor[HTML]{FFFFBB}0.971 & \cellcolor[HTML]{FFFFBB}0.991 & \cellcolor[HTML]{F9DAB7}0.951 & 0.966 & \cellcolor[HTML]{FFFFBB}0.991 & 0.978 & 0.975 & 0.980 & \cellcolor[HTML]{FFFFBB}0.934 & 22.62 & 10,100 \\
Ours-S & \cellcolor[HTML]{FFFFBB}0.971 & 0.986 & 0.936 & \cellcolor[HTML]{FFFFBB}0.974 & 0.990 & \cellcolor[HTML]{FFFFBB}0.983 & \cellcolor[HTML]{FFFFBB}0.976 & \cellcolor[HTML]{FFFFBB}0.989 & 0.931 & 0.95 & 100 \\
Ours-M & \cellcolor[HTML]{F9DAB7}0.979 & \cellcolor[HTML]{F4B5B4}0.992 & \cellcolor[HTML]{F9DAB7}0.951 & \cellcolor[HTML]{F9DAB7}0.981 & \cellcolor[HTML]{F9DAB7}0.993 & \cellcolor[HTML]{F9DAB7}0.991 & \cellcolor[HTML]{F9DAB7}0.987 & \cellcolor[HTML]{F4B5B4}0.993 & \cellcolor[HTML]{F9DAB7}0.945 & 1.41 & 100 \\
Ours-L & \cellcolor[HTML]{F4B5B4}0.981 & \cellcolor[HTML]{F4B5B4}0.992 & \cellcolor[HTML]{F4B5B4}0.954 & \cellcolor[HTML]{F4B5B4}0.986 & \cellcolor[HTML]{F4B5B4}0.994 & \cellcolor[HTML]{F4B5B4}0.993 & \cellcolor[HTML]{F4B5B4}0.988 & \cellcolor[HTML]{F4B5B4}0.993 & \cellcolor[HTML]{F4B5B4}0.951 & 7.16 & 100 \\ 
\bottomrule
\end{tabular}}

 \caption{SSIM quantitative comparison at $\mathbf{400\times400}$ resolution on Realistic Synthetic 360° dataset. \textit{Size} is model size. \textit{Data Size} is the number of the training images.  \colorbox[HTML]{F4B5B4}{value} means first, \colorbox[HTML]{F9DAB7}{value} means second, \colorbox[HTML]{FFFFBB}{value} means third.}

    \label{tab:Comparison with R2L on 360}
\end{table*}

\begin{table*}
\footnotesize
\centering
\tabcolsep=0.22cm
\resizebox{\linewidth}{!}{
\begin{tabular}{l|c|c|c|c|c|c|c|c|c|c}
\toprule
Method & Avg & Chair & Drums & Ficus & Hotdog & Lego & Materials & Mic & Ship & Size(MB) \\ \hline
VQRF~\cite{li2023compressing} & 0.954 & 0.974 & 0.928 & 0.977 & 0.978 & 0.973 & 0.945 & 0.982 & 0.877 & $\overline{1.43}$ \\ \hline
NeRF~\cite{mildenhall2020nerf} & 0.947 & 0.967 & 0.925 & 0.964 & 0.974 & 0.961 & 0.949 & 0.980 & 0.856 & 5.00 \\
Mip-NeRF~\cite{barron2021mip} & 0.961 & \cellcolor[HTML]{FFFFBB}0.981 & \cellcolor[HTML]{FFFFBB}0.932 & \cellcolor[HTML]{FFFFBB}0.980 & \cellcolor[HTML]{F9DAB7}0.982 & 0.978 & \cellcolor[HTML]{FFFFBB}0.959 & \cellcolor[HTML]{F9DAB7}0.991 & 0.882 & 2.50 \\
TensoRF~\cite{chen2022tensorf} & \cellcolor[HTML]{F9DAB7}0.963 & \cellcolor[HTML]{F4B5B4}0.985 & \cellcolor[HTML]{F4B5B4}0.937 & \cellcolor[HTML]{F4B5B4}0.982 & \cellcolor[HTML]{F9DAB7}0.982 & \cellcolor[HTML]{F9DAB7}0.983 & 0.952 & 0.988 & \cellcolor[HTML]{F9DAB7}0.895 & 71.80 \\
I-NGP~\cite{muller2022instant} & \cellcolor[HTML]{F9DAB7}0.963 & 0.979 & \cellcolor[HTML]{F4B5B4}0.937 & \cellcolor[HTML]{F9DAB7}0.981 & \cellcolor[HTML]{F9DAB7}0.982 & \cellcolor[HTML]{FFFFBB}0.982 & 0.951 & \cellcolor[HTML]{FFFFBB}0.990 & \cellcolor[HTML]{F4B5B4}0.896 & 64.10 \\ \hline
Ours-M & 0.956 & 0.973 & 0.928 & 0.975 & \cellcolor[HTML]{F9DAB7}0.982 & 0.979 & \cellcolor[HTML]{F9DAB7}0.972 & \cellcolor[HTML]{FFFFBB}0.990 & 0.870 & \textbf{1.41} \\
Ours-L & \cellcolor[HTML]{F4B5B4}0.964 & \cellcolor[HTML]{F9DAB7}0.982 & \cellcolor[HTML]{FFFFBB}0.932 & \cellcolor[HTML]{FFFFBB}0.980 & \cellcolor[HTML]{F4B5B4}0.984 & \cellcolor[HTML]{F4B5B4}0.984 & \cellcolor[HTML]{F4B5B4}0.975 & \cellcolor[HTML]{F4B5B4}0.992 & \cellcolor[HTML]{FFFFBB}0.886 & 7.16 \\ 
\bottomrule
\end{tabular}}

\caption{SSIM quantitative comparison at $\mathbf{800\times800}$ resolution on Realistic Synthetic 360° dataset. The VQRF is based on DVGO~\cite{sun2022direct}. \textit{Size} is model size. \textit{Data Size} is the number of training images. \textbf{The first part} is NeRF's compression method. \textbf{The second part} is the SoTA NeRF methods. \textbf{The third part} is the SoTA NeLF methods.}

    \label{tab:Comparison with Other Methods on 360}
\end{table*}

\begin{table*}

    \centering
    \tabcolsep=0.05cm
    \resizebox{\linewidth}{!}{
\begin{tabular}{l|cccccc|ccc|c}
\toprule
Method & \multicolumn{6}{c|}{Dense} & \multicolumn{3}{c|}{Sparse} & \begin{tabular}[c]{@{}c@{}}Additional\\ Images\end{tabular} \\ \cline{2-10}
 & \multicolumn{1}{l|}{Avg} & \multicolumn{1}{l|}{Horns(54)} & \multicolumn{1}{l|}{Trex(48)} & \multicolumn{1}{l|}{Room(36)} & \multicolumn{1}{l|}{Fortress(36)} & \multicolumn{1}{l|}{Flower(29)} & \multicolumn{1}{l|}{Leaves(21)} & \multicolumn{1}{l|}{Orchids(21)} & \multicolumn{1}{l|}{Fern(17)} & \multicolumn{1}{l}{} \\ \hline
NeRF~\cite{mildenhall2020nerf} & \multicolumn{1}{c|}{0.917} & \multicolumn{1}{c|}{\cellcolor[HTML]{FFFFBB}0.940} & \multicolumn{1}{c|}{0.947} & \multicolumn{1}{c|}{\cellcolor[HTML]{FFFFBB}0.979} & \multicolumn{1}{c|}{\cellcolor[HTML]{FFFFBB}0.958} & 0.923 & \multicolumn{1}{c|}{0.844} & \multicolumn{1}{c|}{\cellcolor[HTML]{F9DAB7}0.794} & \cellcolor[HTML]{F4B5B4}0.899 & 0 \\
Attal \etal \cite{attal2022learning} & \multicolumn{1}{c|}{\cellcolor[HTML]{F9DAB7}0.956} & \multicolumn{1}{c|}{\cellcolor[HTML]{F9DAB7}0.955} & \multicolumn{1}{c|}{\cellcolor[HTML]{F9DAB7}0.959} & \multicolumn{1}{c|}{\cellcolor[HTML]{FFFFBB}0.979} & \multicolumn{1}{c|}{0.954} & \cellcolor[HTML]{FFFFBB}0.934 & \multicolumn{1}{c|}{\cellcolor[HTML]{FFFFBB}0.847} & \multicolumn{1}{c|}{\cellcolor[HTML]{FFFFBB}0.766} & 0.850 & 0 \\
R2L~\cite{wang2022r2l} & \multicolumn{1}{c|}{\cellcolor[HTML]{FFFFBB}0.950} & \multicolumn{1}{c|}{0.934} & \multicolumn{1}{c|}{\cellcolor[HTML]{FFFFBB}0.948} & \multicolumn{1}{c|}{\cellcolor[HTML]{F4B5B4}0.980} & \multicolumn{1}{c|}{\cellcolor[HTML]{F9DAB7}0.959} & \cellcolor[HTML]{F9DAB7}0.929 & \multicolumn{1}{c|}{\cellcolor[HTML]{F4B5B4}0.860} & \multicolumn{1}{c|}{\cellcolor[HTML]{F4B5B4}0.800} & \cellcolor[HTML]{F9DAB7}0.895 & 10,000 \\
Ours & \multicolumn{1}{c|}{\cellcolor[HTML]{F4B5B4}0.961} & \multicolumn{1}{c|}{\cellcolor[HTML]{F4B5B4}0.961} & \multicolumn{1}{c|}{\cellcolor[HTML]{F4B5B4}0.961} & \multicolumn{1}{c|}{\cellcolor[HTML]{FFFFBB}0.979} & \multicolumn{1}{c|}{\cellcolor[HTML]{F4B5B4}0.963} & \cellcolor[HTML]{F4B5B4}0.939 & \multicolumn{1}{c|}{\cellcolor[HTML]{F9DAB7}0.858} & \multicolumn{1}{c|}{0.716} & \cellcolor[HTML]{FFFFBB}0.873 & 0 \\ \bottomrule
\end{tabular}}
 \caption{SSIM quantitative comparison at $\mathbf{504\times378}$ resolution on Real Forward-Facing. \textit{Additional Images} is the required number of extra synthesized images. The number of original training views is in the brackets.}
    \label{tab:Comparison with R2L on Forward-Facing}
\end{table*}

\begin{table*}
    \centering
    \tabcolsep=0.05cm

    \resizebox{\linewidth}{!}{
    \begin{tabular}{l|cccccc|ccc}
\toprule
Method & \multicolumn{6}{c|}{Dense} & \multicolumn{3}{c}{Sparse} \\ \cline{2-10} 
 & \multicolumn{1}{c|}{Avg} & \multicolumn{1}{c|}{Horns(54)} & \multicolumn{1}{c|}{Trex(48)} & \multicolumn{1}{c|}{Room(36)} & \multicolumn{1}{c|}{Fortress(36)} & Flower(29) & \multicolumn{1}{c|}{Leaves(21)} & \multicolumn{1}{c|}{Orchids(21)} & Fern(17) \\ \hline
NeRF~\cite{mildenhall2020nerf} & \multicolumn{1}{c|}{0.873} & \multicolumn{1}{c|}{0.828} & \multicolumn{1}{c|}{0.880} & \multicolumn{1}{c|}{\cellcolor[HTML]{F4B5B4}0.948} & \multicolumn{1}{c|}{0.881} & 0.827 & \multicolumn{1}{c|}{0.690} & \multicolumn{1}{c|}{0.641} & 0.792 \\
Plenoxels~\cite{fridovich2022plenoxels} & \multicolumn{1}{c|}{\cellcolor[HTML]{FFFFBB}0.886} & \multicolumn{1}{c|}{\cellcolor[HTML]{FFFFBB}0.857} & \multicolumn{1}{c|}{\cellcolor[HTML]{F9DAB7}0.890} & \multicolumn{1}{c|}{0.937} & \multicolumn{1}{c|}{\cellcolor[HTML]{F9DAB7}0.885} & \cellcolor[HTML]{F4B5B4}0.862 & \multicolumn{1}{c|}{\cellcolor[HTML]{F9DAB7}0.760} & \multicolumn{1}{c|}{\cellcolor[HTML]{F9DAB7}0.687} & \cellcolor[HTML]{F9DAB7}0.832 \\
TensoRF~\cite{chen2022tensorf} & \multicolumn{1}{c|}{\cellcolor[HTML]{F4B5B4}0.889} & \multicolumn{1}{c|}{\cellcolor[HTML]{F9DAB7}0.859} & \multicolumn{1}{c|}{\cellcolor[HTML]{F9DAB7}0.890} & \multicolumn{1}{c|}{\cellcolor[HTML]{FFFFBB}0.946} & \multicolumn{1}{c|}{\cellcolor[HTML]{F4B5B4}0.889} & \cellcolor[HTML]{FFFFBB}0.859 & \multicolumn{1}{c|}{\cellcolor[HTML]{FFFFBB}0.746} & \multicolumn{1}{c|}{\cellcolor[HTML]{FFFFBB}0.655} & \cellcolor[HTML]{FFFFBB}0.816 \\ \hline
Ours & \multicolumn{1}{c|}{\cellcolor[HTML]{F4B5B4}0.889} & \multicolumn{1}{c|}{\cellcolor[HTML]{F4B5B4}0.862} & \multicolumn{1}{c|}{\cellcolor[HTML]{F4B5B4}0.897} & \multicolumn{1}{c|}{\cellcolor[HTML]{F4B5B4}0.948} & \multicolumn{1}{c|}{0.881} & 0.856 & \multicolumn{1}{c|}{0.713} & \multicolumn{1}{c|}{0.547} & 0.735 \\ \bottomrule
\end{tabular}}

\caption{SSIM quantitative comparison at $\mathbf{1008\times756}$ resolution on Real Forward-Facing. \textit{Additional Images} is the required number of extra synthesized images. The number of original training views is in the brackets.}

    \label{tab:Comparison with R2L on Forward-Facing}
\end{table*}

\end{document}